\documentclass[
]{ceurart}

\usepackage{listings}
\usepackage[linesnumbered,ruled,vlined]{algorithm2e}
\usepackage{natbib}
\begin{document}

\copyrightyear{2022}
\copyrightclause{Copyright for this paper by its authors.
  Use permitted under Creative Commons License Attribution 4.0
  International (CC BY 4.0).}

\conference{CLEF 2022: Conference and Labs of the Evaluation Forum, 
    September 5--8, 2022, Bologna, Italy}


\title{Solutions for Fine-grained and Long-tailed Snake Species Recognition in SnakeCLEF 2022}

\author[1]{Cheng Zou}[%
email=wuyou.zc@antgroup.com,
]
\address[1]{Ant Group,
  Yuan Space 556 Xixi Road, Hangzhou, 310013, China}

\author[1]{Furong Xu}[%
email=booyoungxu.xfr@antgroup.com,
]

\author[1]{Meng Wang}[%
email=darren.wm@antgroup.com,
]

\author[1]{Wen Li}[%
email=yinian.lw@alibaba-inc.com,
]

\author[1]{Yuan Cheng}[%
email=chengyuan.c@antgroup.com,
]

\begin{abstract}
%   A clear and well-documented \LaTeX{} document is presented as an
%   article formatted for publication by CEUR-WS in a conference
%   proceedings. Based on the ``ceurart'' document class, this article
%   presents and explains many of the common variations, as well as many
%   of the formatting elements an author may use in the preparation of
%   the documentation of their work.
Automatic snake species recognition is important because it has vast potential to help lower deaths and disabilities caused by snakebites. We introduce our solution in SnakeCLEF 2022 for fine-grained snake species recognition on a heavy long-tailed class distribution. First, a network architecture is designed to extract and fuse features from multiple modalities, i.e. photograph from visual modality and geographic locality information from language modality. Then, logit adjustment based methods are studied to relieve the impact caused by the severe class imbalance. Next, a combination of supervised and self-supervised learning method is proposed to make full use of the dataset, including both labeled training data and unlabeled testing data. Finally, post processing strategies, such as multi-scale and multi-crop test-time-augmentation, location filtering and model ensemble, are employed for better performance. With an ensemble of several different models, a private score 82.65\%, ranking the 3rd, is achieved on the final leaderboard.
\end{abstract}


\maketitle

\section{Introduction}

Snakebite is a global health problem, especially in remote geographic areas and developing countries. According to~\cite{borsodi2021incorporation}, in Asia, up to two million people are envenomed by
snakes each year, while in Africa, there are about 435,000 to 580,000 snakebites annually
that need treatment, for they can cause permanent disability and disfigurement. Taxonomic knowledge about snakes is crucial in diagnosis and medical response to snakebites, and accurate identification of the snake species is important for the appropriate treatment of
snakebite victims since specific antivenoms are effective against specific venomous snakes~\cite{chamidullin2021deep}. Manual identification, e.g. training doctors on each species, is no easy feat, because there are more than 3,500 species of snakes, 600 of which are venomous~\cite{coca2021uaic}. So, building an automatic and robust image-based system for snake species identification has the greatest potential to save lives~\cite{picek2021overview}.

% Building an automatic and robust image-based system for snake species identification is an important goal for biodiversity, conservation, and global health~\cite{PicekEtAl:CLEF-2021}. 
% With recent estimates of 81,410–137,880 deaths and up to three times as many victims of amputations, permanent disability and disfigurement (globally each year) caused by venomous snakebite, such a system has the potential to improve eco-epidemiological data and treatment outcomes. 
% This applies especially in remote geographic areas and developing countries, where automatic snake species identification has the greatest potential to save lives. 

The difficulty of snake species identification, from both a human and a machine perspective, lies in the high intra-class and low inter-class variance in appearance, which may depend on geographic location, colour morph, sex, or age~\cite{picek2021overview}. Sometimes, having the image alone is not enough, because many species are visually similar to other species, while introducing the geographic origin of an observation can help to recognize considerably.
% In no location on Earth do more than 126 of the approximately 3,900 snake species co-occur. Thus, regularization to all countries is a critical component of any snake identification method.
The task of SnakeCLEF 2022 challenge~\cite{snakeclef2022}, as part of the LifeCLEF 2022~\cite{lifeclef2022,joly2022lifeclef}, aims to recognize a snake species ID given multiple photographs of the same individual and its corresponding geographic locality information.

In this paper, we introduce the solution of team "SAI" in SnakeCLEF 2022 for fine-grained snake species recognition on a severe long-tailed class distribution. First, as discussed above, using the image data alone is not enough, more cues are required for prediction, so a network architecture is designed to extract and fuse features from multiple modalities, i.e. photograph from visual modality and geographic locality information from language modality. Then, because of the heavy long-tailed class distribution, some logit adjustment based methods are studied to relieve the impact caused by the severe class imbalance, which significantly improves the performance. Next, to make full use of the dataset, including both labeled training data and unlabeled testing data, a combination of supervised and self-supervised learning method is utilized for pretraining. Finally, some post processing strategies are employed for better performance, including multi-scale and multi-crop test-time-augmentation, location filtering and model ensemble.

\section{Related Work}
\subsection{Fine-grained Vision Classification}
Snake species recognition is basically a task of fine-grained vision classification (FGVC). Modern fine-grained image classification methods can be divided into two parts, use image data only~\cite{lin2015bilinear, zheng2019learning, he2021transfg, ge2019weakly}, and use image data as well as extra data from other modalities~\cite{chu2019geo, mac2019presence, he2017fine, diao2022metaformer}. For the task of snake species recognition, one can solve it by using image data only, but a better choice is to use both image data and geographic locality information. In such studies with extra modalities, ~\cite{chu2019geo} is a classic method to introduce geographic information, ~\cite{mac2019presence} introduced spatio-temporal information into the network. MetaFormer~\cite{diao2022metaformer} proposed a unified and flexibly framework to joint the visual appearance and various meta-information.

\subsection{Long-tailed Distribution}
Snake species recognition in real world is also a task of long-tailed image classification. Recently, long-tailed learning has received plenty of research interests, and among which there are mainly two kinds of solutions, one is post-hoc normalisation of the classification weights~\cite{zhang2019balance, kim2020adjusting, kang2019decoupling, ye2020identifying}, and the other is modification of the loss to account for varying class penalties~\cite{tan2020equalization, cao2019learning, cui2019class}. ~\cite{menon2020long} revisited the classic idea of logit adjustment based on the label frequencies, either applied post-hoc to a trained model, or enforced in the loss during training.

\subsection{Snake Species Classification}
Before us, there have been a few works tried to build automatic snake species recognition systems. In~\cite{borsodi2021incorporation,bloch2021efficientnets}, object detectors were first trained to reduce clutter and drop the unnecessary background for preprocessing, and then the detected snakes were classified by trained deep models. \cite{desingu2021snake} extracted feature for each image with Inception ResNet V2 and concatenated it with geographic feature, then the concatenated features were classified using a lightweight gradient boost classifier. In~\cite{bloch2021efficientnets}, EfficientNet and Vision Transformer were trained, and the prior probabilities of location information were multiplied with the model predictions in a subsequent step. Besides, \cite{chamidullin2021deep,coca2021uaic,bloch2021efficientnets,kalinathan2021automatic} used multiple models to improve the performance.

\section{Methodology}
\subsection{Overview}
The proposed solution for fine-grained snake species recognition on a long-tailed class distribution mainly consists of four parts: 1) a network architecture to extract and fuse features from multiple modalities, 2) logit adjustment to relieve the impact caused by the severe class imbalance. 3) a combination of supervised and self-supervised learning method to make full use of both labeled training data and unlabeled testing data, 4) post processing strategies such as multi-scale and multi-crop test-time-augmentation, location filtering and model ensemble.

\begin{figure*}
	\begin{center}
		\includegraphics[width=1.0\linewidth]{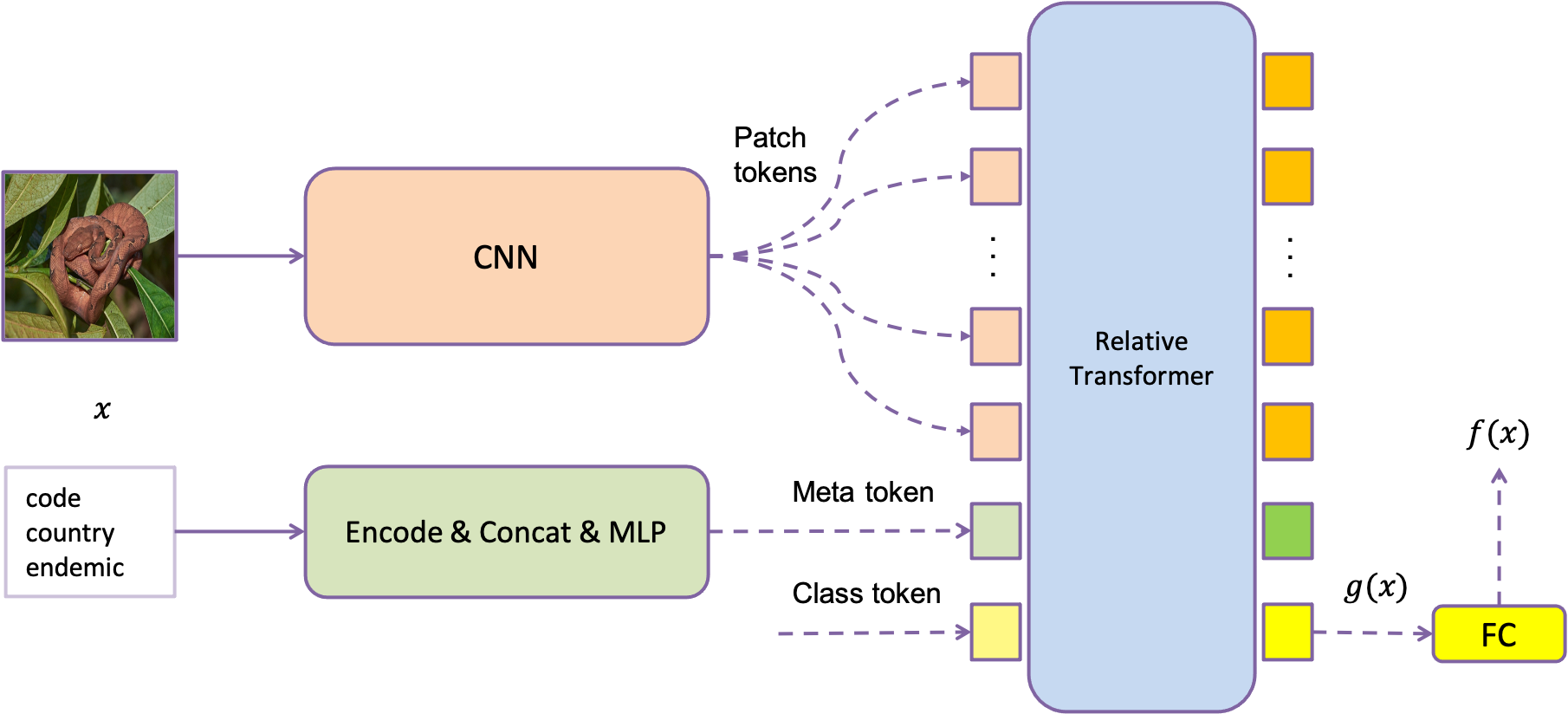}
	\end{center}
	\caption{Overall network architecture. It has a hybrid framework, where CNN branch is used to extract vision features, MLP branch is used to encode meta data, and transformer is used to fuse vision features and meta information. $g(x)$ stands for extracted features and $f(x)$ stands for the predicted logits.}
	\label{fig:framework}
\end{figure*}

\subsection{Network Architecture}
The network design follows MetaFormer~\cite{diao2022metaformer}. It has a hybrid framework, where on one branch CNN is used to extract vision features, on the other branch MLP is used to encode meta data, and transformer is later used to fuse vision features and meta information. The output feature map of CNN branch is transformed to a series of patch embeddings (denoted as patch tokens), along with the output embedding of MLP branch (denoted as meta token), along with the class token, are fed into the transformer layers for prediction. In the task of snake species recognition, the meta data available consists of location code, country and the tag of endemic.

\subsection{Logit Adjustment for Long-tailed Learning}
According to statistics, the dataset has a heavy long-tailed class distribution, where the most frequent species is represented by 6,472 images and the least frequent species by just 5 samples. It is also noteworthy that the evaluation metric, the Mean (Macro) F1-Score,  weights recall and precision equally, and a good retrieval algorithm will maximize both precision and recall simultaneously. Thus, moderately good performance on both will be favoured over extremely good performance on one and poor performance on the other. If there is no action to deal with long-tailed distribution problem, the recall for rare classes would be very low, thus the overall performance could not be high.

Logit adjustment~\cite{menon2020long} is adopted to relieve the long-tailed distribution problem. In post-hoc adjustment, we predict the following instead of the original one,
\begin{equation}
    {\rm {argmax}}_{y\in[L]} \exp(f(x)_{y})/\pi_{y}^{\tau}={\rm {argmax}}_{y\in[L]} f(x)_{y}-\tau \cdot \log\pi_{y}
    \label{eq-post-hoc}
\end{equation}

In logit adjusted loss, it is combined with the soft target cross entropy loss, 
\begin{equation}
    L(f(x)_{y}, y) = -y \cdot \log {\rm softmax}(f(x)_{y}+\tau \cdot \log\pi_{y})
\end{equation}

where $\tau > 0$ is a hyper-parameter, $f(x)_{y}$ is the output logits of the neural network given input $x$, and $\pi_{y}$ is the estimate of the class prior, e.g., the empirical class frequency on the training data. In practice, one can use either post-hoc logit adjustment or logit adjusted loss.

\subsection{Combination of Supervised and Self-supervised Learning}
In order to make full use of the dataset, including both labeled training data and unlabeled testing data, a combination of supervised and self-supervised training framework is proposed for pretraining. Specifically, we perform supervised learning on labelled training data and self-supervised learning (SSL) on all the data to obtain a set of task-related parameter initialization. Since the task provides images and meta information, we do pretraining for MetaFormer~\cite{diao2022metaformer} with meta data, which can jointly take advantage of vision and meta-information. 

To obtain a task-relevant initialization instead of imagenet initialization, we combine the self-supervised method SimCLR~\cite{chen2020simple} with MetaFormer. Specifically, for each input pair of image and meta data, we randomly perform two data augmentations (Fig.~\ref{fig:augmentation}) for the image, but no augmentation on meta data, then a classification loss SoftTargetCE~\cite{muller2019does,zhang2021delving} (short for soft target cross entropy) is applied to the labeled data only, and a contrastive loss InfoNCE~\cite{oord2018representation} is applied to all the data. Thus, the loss function for pretraining is,
\begin{equation}
    L_{pretrain}(X, Y)={\rm SoftTargetCE}_{Y\neq-1}(f(X), Y) + \alpha \cdot {\rm {InfoNCE}}(g([X;\bar{X}]), Y)
    \label{eq-pretrain}
\end{equation}

where $f(X)$ stands for logits, $g(X)$ stands for extracted features for a given batch $X$, and $Y$ stands for its corresponding label, $Y=-1$ means unlabeled testing data. $\bar{X}$ is an augmented version of $X$. $\alpha$ is a hyper-parameter to balance the relative importance between supervised loss and self-supervised loss.

\subsection{Post Processing}
Two kinds of post processing strategies are mainly used, one is multi-scale, multi-crop and multi-model ensemble, the other is location filtering. For model ensemble, average logits is first calculated based on the output logits of different models, then the mean logits is adjusted to deal with long-tailed distribution problem. Specifically, for each single scale/crop input $x_i$, the $j\rm {-th}$ model outputs logits $f_{j}(x_i)$, then the final logits used for prediction is,
\begin{equation}
    {\rm logits\_adjusted} = {\rm mean}(\sum_{j} \sum_{i} f_{j}(x_i)) - \tau \cdot \log\pi
\end{equation}

\noindent\textbf{Location Filtering.} During training, it can be found that top-5 accuracy is much higher than that of top-1, which implies that if properly chosen from top predictions, the result could be better than the naive argmax one. A locations-to-species mapping is used to heuristically choose the best candidate prediction. For simplicity, we iterate through a sorted logits list until the first certain species appears, whose species name co-occurs with its location code. Algorithm~\ref{Location_filtering} shows the numpy style pseudo code.
\begin{algorithm}[t]
    \SetKwInOut{Input}{Input}
    \SetKwInOut{Output}{Output}

	\Input{logits\_adjusted, locations-to-species mapping $L2S$, meta data $Meta$}
	\Output{predicted species name}
	\BlankLine 

    idx\_sort = np.argsort(logits\_adjusted)[::-1]\

    \For{\rm idx in idx\_sort}{
        species\_name = species\_list[idx]\

        \If()
            {{\rm species\_name in} L2S{\rm[}Meta{\rm[`code']]}}{break}
    }

    return species\_name

 	\caption{Location filtering}
 	\label{Location_filtering} 
\end{algorithm}

\section{Experiments}
\subsection{Experimental Settings}
\noindent\textbf{Dataset}: The dataset is based on 187,129 snake observations with 318,532 photographs belonging to 1,572 snake species and observed in 208 countries. The data were gathered from the online biodiversity platform iNaturalist. The provided dataset has a heavy long-tailed class distribution, where the most frequent species (Natrix natrix) is represented by 6,472 images and the least frequent species by just 5 samples.

\begin{figure*}
	\begin{center}
		\includegraphics[width=1.0\linewidth]{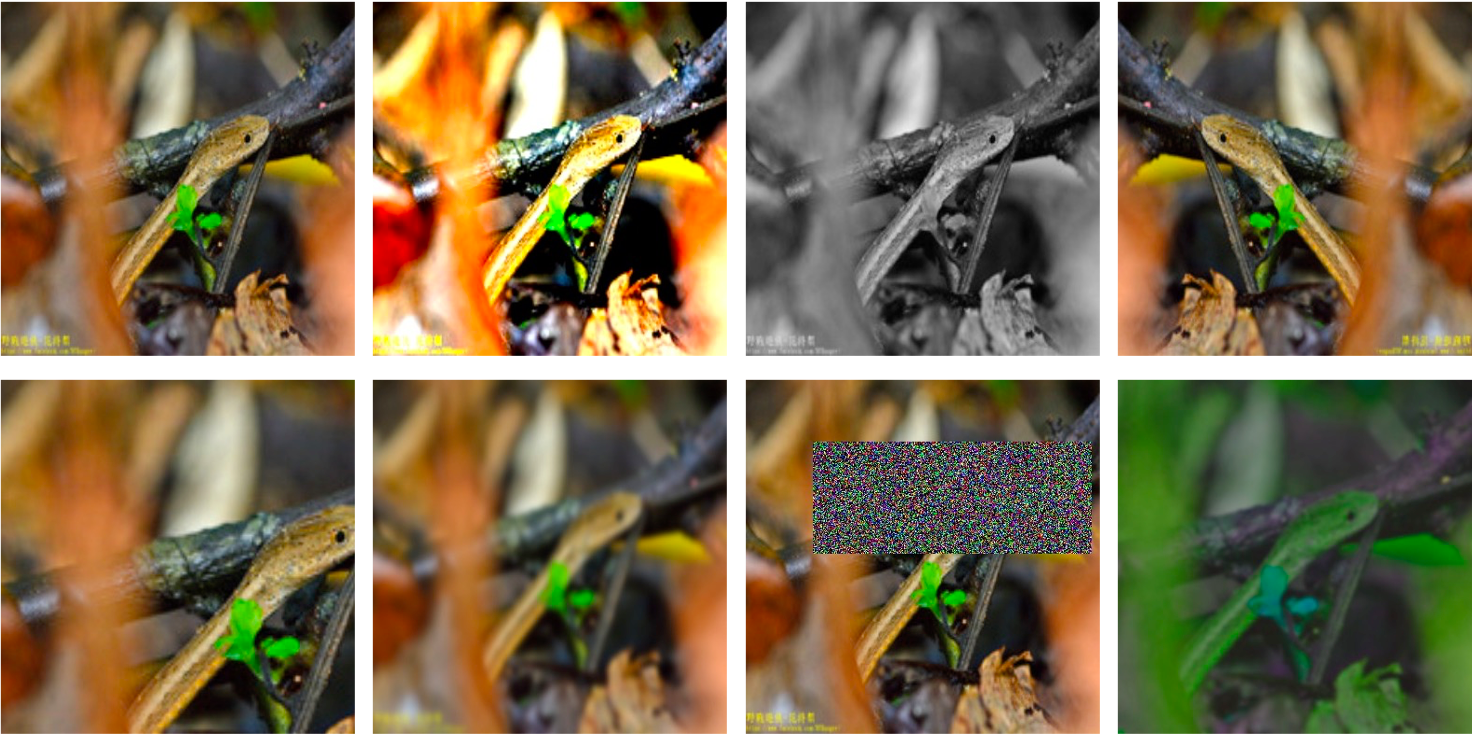}
	\end{center}
	\caption{Augmentations for self-supervised learning. From left to right, top to bottom: original image without augmentation, color jittering, gray scale, horizontal flip, random resized crop, Gaussian blur, random erasing, a composed random augmentation of the above.}
	\label{fig:augmentation}
\end{figure*}

\begin{table}[t]
\setlength{\tabcolsep}{1.6pt}
\caption{Representative experimental results about the importance of: 1) meta information, 2) different pretrained models, 3) solving long-tailed class distribution problem, 4) location filtering, 5) different input image size, 6) model ensemble. It's noteworthy that different parts in the table have different baselines, which is mainly caused by limited number of submissions.}
\label{tab:result}
\begin{tabular}{llccc}
\hline \hline
Experimental Settings & macro $F_1$ & Comments \\ 

\hline
\textit{The importance of meta information} \\
Vision Only & 66.64~\% & ImageNet-21k pretraining + Location filtering \\
Vision + Meta-208d & 71.11~\% & ImageNet-21k pretraining + Location filtering \\
Vision + Meta-2438d & 74.19~\% & ImageNet-21k pretraining + Location filtering \\

\hline
\textit{Different pretrained models} \\
ImageNet-1k~\cite{diao2022metaformer} & 63.76~\% & Without location filtering or logit adjustment \\
ImageNet-21k~\cite{diao2022metaformer} & 66.64~\% & Without location filtering or logit adjustment \\
iNat21~\cite{diao2022metaformer} & 64.32~\% & Without location filtering or logit adjustment \\
Supervised+SSL & 68.83~\% & Without location filtering or logit adjustment \\

\hline
\textit{Solving long-tailed class distribution} \\
None & 74.19~\% & ImageNet-21k pretraining + Location filtering \\
Seesaw~\cite{wang2021seesaw} & 76.49~\% & ImageNet-21k pretraining + Location filtering \\
Logit adjusted loss~\cite{menon2020long} & 78.01~\% & ImageNet-21k pretraining + Location filtering \\
Post-hoc logit adjustment~\cite{menon2020long} & 78.57~\% & ImageNet-21k pretraining + Location filtering \\

\hline
\textit{Location filtering} \\
Without & 64.32~\% & iNat21 + No meta data + No logit adjustment \\
With & 69.09~\% & iNat21 + No meta data + No logit adjustment \\

\hline
\textit{Different input image size} \\
384 & 81.18~\% & Meta+Logit adjustment+SSL+Location filtering \\ 
512 & 82.03~\% & Meta+Logit adjustment+SSL+Location filtering \\ 

\hline
\textit{Model ensemble} \\
Model1 & 81.18~\% & input image size 384 \\ 
Model2 & 82.03~\% & input image size 512 \\
Ensemble & 82.72~\% & \\ 

\hline \hline
\end{tabular}
\end{table}

\noindent\textbf{Evaluation Metric}:
The evaluation metric for this competition is macro $F_1$-Score. The $F_1$ score for the $i$-th species is computed as,
\begin{equation}
    F_1^i=2 \cdot \frac{precision_i \cdot recall_i}{precision_i + recall_i}
\end{equation}

The macro $F_1$ is calculated by averaging the $F_1$ scores over all the species~\cite{borsodi2021incorporation}, 
\begin{equation}
    {\rm macro} ~F_1=\sum_{i=1}^{N} \frac{F_1^i}{N}
\end{equation}

where $N$ is the number of species. The macro $F_1$ score is not biased by class frequencies and is more suitable for the long-tailed class distributions. The $F_1$ metric weights recall and precision equally, and a good retrieval algorithm will maximize both precision and recall simultaneously. Thus, moderately good performance on both will be favoured over extremely good performance on one and poor performance on the other.

\noindent\textbf{Implementation Details}:
MetaFormer~\cite{diao2022metaformer} is selected as our base network. More specifically, we use MetaFormer-2 with extra meta information as input for both pretraining and finetuning. The meta data available for snake species consists of location code, country and the tag of endemic. Thus we construct a 2438-d vector to encode all of the above meta data, and then send it to an MLP to generate meta token. Hyper-parameter $\tau$ in Eq.~\ref{eq-post-hoc} is set to 0.55, $\alpha$ in Eq.~\ref{eq-pretrain} is set to 0.001. In supervised and self-supervised pretraining, ImageNet-21k pretrained model is loaded as initialization, and the learning rate is initialized as $5\times10^{-5}$. In later finetuning, the learning rate is initialized as $5\times10^{-6}$. The models are trained on 8-NVIDIA A100-GPU machines for 300 epochs with a per GPU batch size of 72 for size 384 and 32 for size 512. The augmentations for self-supervised learning is illustrated in Fig.~\ref{fig:augmentation}, while those for supervised learning and finetuning still follows MetaFormer.

\subsection{Experimental Results}
In this part, we report some representative experimental results, including: 1) the importance of meta information, 2) the importance of pretrained models, 3) the importance of solving long-tailed class distribution, 4) the importance of location filtering, 5) the importance of larger input size, 6) the importance of model ensemble.

The experimental results are summarized in Tab.~\ref{tab:result}. Academically, these comparisons here are not strict ablation studies, because they are obtained by few limited submissions during the competition. However, these results provided a meaningful and effective path to optimize the solution, which did improve the online judge performance. 

As shown in Tab.~\ref{tab:result}, training the model with additional meta information significantly improves the performance from 66.64\% to 74.19\%, which indicates the importance of data from multiple modalities. For pretrained models, the proposed task-related supervised+SSL pretraining for Snake performs better than those commonly used ones, which indicates the importance of using unlabeled testing data. In long-tailed learning, logit adjustment is proved to be more effective, which improves the score from 74.19\% to 78.57\%. Location filtering, a task-specific post processing, using the statistics prior from the whole dataset to remove illegal predictions, improves the score from 64.32\% to 69.09\%. Also, training with larger input size 512 improves the performance from 81.18\% to 82.03\%.

With an ensemble of seven different models, we got private score 82.65\% on the final leaderboard. The improvement is marginal compared to a single model, because the differences among these models are small, i.e., different epochs, different hyper-parameters. Interestingly, in late submission, we find it inferior to 82.72\%, the ensemble of only two models.

\section{Conclusion and Future Work}
In this paper, we introduce our solution in SnakeCLEF 2022 for fine-grained snake species recognition on a severe long-tailed class distribution. Attentions have been mainly focused on four parts: 1) fusion of vision features and meta information, 2) solving long-tailed class distribution problem, 3) making full use of both the labeled training data and unlabeled testing data via supervised and self-supervised pretraining, 4) post processing strategies such as location filtering and model ensemble. 

Though great improvements have been made, there still exist some actions of great potential for future work: 1) hard example mining for fine-grained and long-tailed dataset, 2) treating it as a retrieval task not a classification task, 3) using a snake detection model to get more precise bounding box for data pre-processing. We have tried some of the above but none of them contributed to the final performance during the competition, but they have great potential if further studied.

%%
%% Define the bibliography file to be used
\bibliography{main}

\appendix

\section{Online Resources}

The code and models are available at \href{https://drive.google.com/drive/folders/1_I2yjUQah99kz0PyPICyp6STjpS809cl?usp=sharing}{Google Drive}.

% \section{Team Merges}

% Our team was not merged at that time due to some reasons. Here is a team merge request, or just a clarification. Our nice and hard working team members are `SAI', `booyoungxu', `Dafeilong' and `jinyani'.

\end{document}